\def\year{2020}\relax
\documentclass[letterpaper]{article} 
\usepackage{array}
\usepackage{booktabs}
\usepackage{subfigure}
\usepackage{algorithm}
\usepackage{algorithmicx}
\usepackage[noend]{algpseudocode}
\usepackage{amsmath}
\usepackage{amsfonts}
\usepackage{amssymb}
\usepackage{color}
\usepackage{aaai20}  
\usepackage{times}  
\usepackage{helvet} 
\usepackage{courier}  
\usepackage[hyphens]{url}  
\usepackage{graphicx} 
\usepackage{graphicx}  
\newcommand\first{\textbf{WS-UDA}}
\newcommand\second{\textbf{2ST-UDA}}
\title{Adversarial Training Based Multi-Source Unsupervised Domain Adaptation\\ for Sentiment Analysis }
\author{Yong Dai\textsuperscript{\rm 1}, Jian Liu\textsuperscript{\rm 1}, Xiancong Ren\textsuperscript{\rm 1}, Zenglin Xu\textsuperscript{\rm 2,\rm 3}\\ 
\textsuperscript{\rm 1}SMILE Lab, School of Computer Science and Engineering, University of Electronic Science and Technology of China\\
Chengdu, Sichuan, China\\
\textsuperscript{\rm 2}School of Computer Science and Technology, Harbin Institute of Technology Shenzhen, Shenzhen, Guangdong, China\\
\textsuperscript{\rm 3}Center for Artificial Intelligence, Peng Cheng Laboratory, Shenzhen, Guangdong, China\\
daiyongya@yahoo.com, \{liujian.dl, xiancongRen\}@std.uestc.edu.cn, xuzenglin@hit.edu.cn 
}
 
\begin{document}

\maketitle

\begin{abstract}
Multi-source unsupervised domain adaptation (MS-UDA) for sentiment analysis (SA) aims to leverage useful information in multiple source domains to help do SA in an unlabeled target domain that has no supervised information. Existing algorithms of MS-UDA either only exploit the shared features, i.e., the domain-invariant information, or based on some weak assumption in NLP, e.g., smoothness assumption. To avoid these problems, we propose two transfer learning frameworks based on the multi-source domain adaptation methodology for SA by combining the source hypotheses to derive a good target hypothesis. The key feature of the first framework is a novel Weighting Scheme based Unsupervised Domain Adaptation framework (\first), which combine the source classifiers to acquire pseudo labels for target instances directly. While the second framework is a Two-Stage Training based Unsupervised Domain Adaptation framework (\second), which further exploits these pseudo labels to train a target private extractor. Importantly, the weights assigned to each source classifier are based on the relations between target instances and source domains, which measured by a discriminator through the adversarial training. Furthermore, through the same discriminator, we also fulfill the separation of shared features and private features. Experimental results on two SA datasets demonstrate the promising performance of our frameworks, which outperforms unsupervised state-of-the-art competitors.
\end{abstract}

\section{Introduction}
Sentiment analysis (SA) is a computational study of humans' opinions, sentiments, attitudes towards products, services, etc. Due to the strong theoretical and practical exploring value, SA has been a research hotspot from the early time of this century. For
example, the understanding of customers' emotional tendency is a key to  reshaping business (e.g., the market system of Amazon).
Naturally, there are different opinions commented on diverse kinds of products or services and they can be regarded as located in different domains. Because of the domain discrepancy, we often suppose these diverse opinions come from different distributions and have different characteristics. For illustration, we consider the following two reviews of three products, i.e., \emph{Phones}, \emph{Battery}, and \emph{Car} from Amazon.com. 
\begin{center}
(1) \emph{It looks good.} \bigskip
(2) \emph{It runs fast.}
\vspace{-0.5cm}
\end{center}
Apparently, the first sentence is positive for all products, while the second sentence is positive for \emph{Car} and \emph{Phones}, but negative for \emph{Battery}. Intuitively, there are domain-invariant and domain-specific characteristics between different domains. For a specific domain, supervised learning algorithms have been successfully explored to build sentiment classifiers based on the positive or negative labels~\cite{socher2013recursive,liu2015sentiment}. However, there exists insufficient or no labeled data in a target domain of interest in actual scenarios, while labeling data in this domain may be time-consuming and expensive. Therefore, cross-domain sentiment analysis, which borrows knowledge from related source domains with abundant labeled data to improve the target domain, has become a promising direction. Specially, cross-domain SA in the multi-source unsupervised domain adaptation (MS-UDA) setting, where there are multiple source domains available together with one unlabeled target domain, is more practical and challenging.

To better deal with problems in the MS-UDA setting, researchers have proposed well-developed algorithms. 
We can simply categorize them into two groups \cite{sun2015survey}. The first group of approaches is based on feature representation, which aims to make the feature distributions of source and target domain similar, either by penalizing or removing features whose statistics vary between domains or by learning a feature space embedding or projection in which a distribution divergence statistic is minimized. For example, \citeauthor{liu2017adversarial} (\citeyear{liu2017adversarial}) firstly introduce adversarial training and orthogonality constraints to derive more pure shared features for simultaneously handling multiple domains of text classification. \citeauthor{chen2018multinomial} (\citeyear{chen2018multinomial}) extend this model and provide theoretical guarantees. However, these methods suffer from the loss of private knowledge when applied in the unsupervised setting. Another group of methods seeks to assign a weight for each pre-learned classifier according to the relationship between the source domain and the target domain. \citeauthor{chattopadhyay2012multisource,sun2011two} (\citeyear{chattopadhyay2012multisource,sun2011two}) assign different weights to different source classifiers based on smoothness assumption. Nevertheless, the smoothness assumption is not always true in NLP. For instance, the embeddings from~\cite{mikolov2013efficient} tell that the word 'like' is most close to the 'unlike', but the sentiment polarity is opposite, which is inconsistent with the smoothness assumption. Besides, the performance of these methods often decreases obviously when directly applied in the unsupervised setting, which is seldom specially studied in the age of neural networks~\cite{zhao2018multiple}.
 
In this paper, we focus on the MS-UDA for SA and desire to combine the hypotheses of multiple labeled source domains to derive a good hypothesis for an unlabeled target domain. For this purpose, we introduce two transfer learning frameworks.
The first framework is Weighting Scheme based Unsupervised Domain Adaptation (\first), in which we integrate the source classifiers to annotate pseudo labels for target instances directly. Our second framework is a Two-Stage Training based Unsupervised Domain Adaptation method (\second), which further utilize pseudo labels to train a target-specific extractor. The key features of our frameworks include: Firstly, we induce the data-dependent prior to our model by considering a discriminator as a probability distribution estimator. Concretely, we exploit the discriminator to measure the instance-to-domain relations between different source domains and target instances, based on which we implement instance-level weighting scheme to assign different weight for each source classifier; Secondly, our frameworks explicitly model both private and shared components of the domain representations and encourage them to be separated or independent, which can resist the contamination by noise that is correlated with the underlying shared distribution~\cite{salzmann2010factorized} and beneficial for system performance~\cite{bousmalis2016domain}. In detail, our frameworks force the shared features to be domain invariant and private features to contain domain-specific information through adversarial training other than the orthogonality constraint adopted by \citeauthor{bousmalis2016domain,liu2017adversarial} (\citeyear{bousmalis2016domain,liu2017adversarial}).


Our contributions are summarized as follows:
\begin{itemize}
    \item We propose two end-to-end frameworks to implement multi-source unsupervised domain adaptation for sentiment analysis by combining multiple source hypotheses. The difference between the two frameworks is how to utilize the pseudo labels annotated by source classifiers. Specially, we regard a discriminator as the metric tool for measuring the instance-to-domain relations and providing different weights to source classifiers. Moreover, the separation of shared features and private features is also realized through adversarial training.
    \item Empirically the proposed frameworks can significantly outperform the state-of-the-art methods.
\end{itemize}
\section{Related works}
\subsubsection{Sentiment analysis} 
SA is commonly regarded	as a special case of classification. One key point of document-level SA is how to represent a document. Traditional machine learning methods represent documents as a bag of its words~\cite{moraes2013document}, in which the word order is ignored and they barely encode the semantics of the words. 
Based on word embedding techniques~\cite{le2014distributed}, many architectures have been explored, such as CNN~\cite{johnson2014effective}, LSTM~\cite{chen2016neural}, Attention-LSTM~\cite{zhou2016attention}. 
In our paper, we also adopt pre-trained embedding and employ an effective multi-task model to conduct SA in the MS-UDA setting. Importantly, the architectures in our frameworks are relatively flexible and can be adapted by the practitioners to suit particular classification tasks.

\subsubsection{Multi-source domain adaptation} 
Multi-source cross-domain SA has raised much attention in recent years~\cite{sun2015survey}. 
The most common way for dealing with multi-source data is to consider all the source domains as one source, which ignores the difference among the sources. The second way is to train per source separately and to  combine multiple base classifiers. For example, \citeauthor{chattopadhyay2012multisource}~(\citeyear{chattopadhyay2012multisource}) introduced a framework, which assigned different weights to different source domains based on the smoothness assumption. \citeauthor{sun2011two}~(\citeyear{sun2011two}) further extended this method and presented a two-stage domain adaptation methodology. \citeauthor{duan2009domain,duan2012domain}~(\citeyear{duan2009domain,duan2012domain}) introduced a data-dependent regularizer into the objective of SVR (support vector regression). The work by \citeauthor{chattopadhyay2012multisource}~(\citeyear{chattopadhyay2012multisource}) is most similar with our framework, and the main differences include: firstly, we exploit an elaborately designed discriminator to help implement the weighting scheme, and secondly, we specialize in an unsupervised setting.
\subsubsection{Adversarial training} 
Adversarial training was adopted to help learn a feature extractor that can map both the source and target input to the same feature space and let the classifier learned on the source data can transfer to the target domain~\cite{ganin2014unsupervised,ganin2016domain,FangXu2018Dart}.
In detail, a classifier was adopted as the domain  discriminator and  the minimax optimization was implemented by using a gradient reverse layer (GRL). \citeauthor{bousmalis2016domain} (\citeyear{bousmalis2016domain}) extended this method and add another reconstruction loss and orthogonality constraint for further improvement. In addition, adversarial training was introduced into multi-task learning in \cite{liu2017adversarial,chen2018multinomial}.
Unlike previous methods just utilizing the adversarial training to obtain the shared features, we employ it to squeeze the domain-related information to the private features and regard it as a probability distribution estimator to derive the instance-to-domain relations.

\section{Multi-source unsupervised domain adaptation }
The goal of our two frameworks is to learn a multi-task model for subjects in all source domains and then integrate the hypothesis generated by each of these source models based on some similarity measures between source domains and target instances. Because of the adoption of the effective multi-task model, our frameworks can exploit the interaction among multiple sources.

In this section, we will introduce two proposed frameworks for multi-source cross-domain sentiment classification. We first present the problem definition and notations, followed by an overview of each framework. Then we detail the frameworks with all components successively.

\subsection{Problem Definition and Notations}
Assume we have access to class labeled (i.e. sentiment polarity) training data from $K$ source domains: $\left \{ \mathcal{S} _j\right \}_{j=1}^{K}$ where $\mathcal{S}_j  \triangleq \left \{ (\boldsymbol{x}_{i}^{\mathcal{S}_j},\boldsymbol{y}_{i}^{\mathcal{S}_j}) \right \}_{i=1}^{\left | \mathcal{S}_j \right |}$, class unlabeled data from a target domain: $\mathcal{T} \triangleq \left \{ \boldsymbol{x}_{i}^{\mathcal{T}} \right \}_{i=1}^{\left | \mathcal{T} \right |}$. In addition, we denote $N_s=\sum_{j=1}^{K}\left | S_j \right |$, $N_\mathcal{T}=\left | \mathcal{T} \right |$ as the number of all labeled source data and unlabeled target data respectively,  and $\mathcal{U} \triangleq \left \{ (\boldsymbol{x}_{i},\boldsymbol{d}_{i}) \right \}_{i=1}^{N}$ as all the domain labeled data, where $N=N_s+N_\mathcal{T}$.
 We can see that all the data regardless of from the target or source domains have domain labels and only the data from source domains have class labels.
\subsection{ Overview of \first}
\begin{figure}[t]
  \centering
  \includegraphics[scale=0.12]{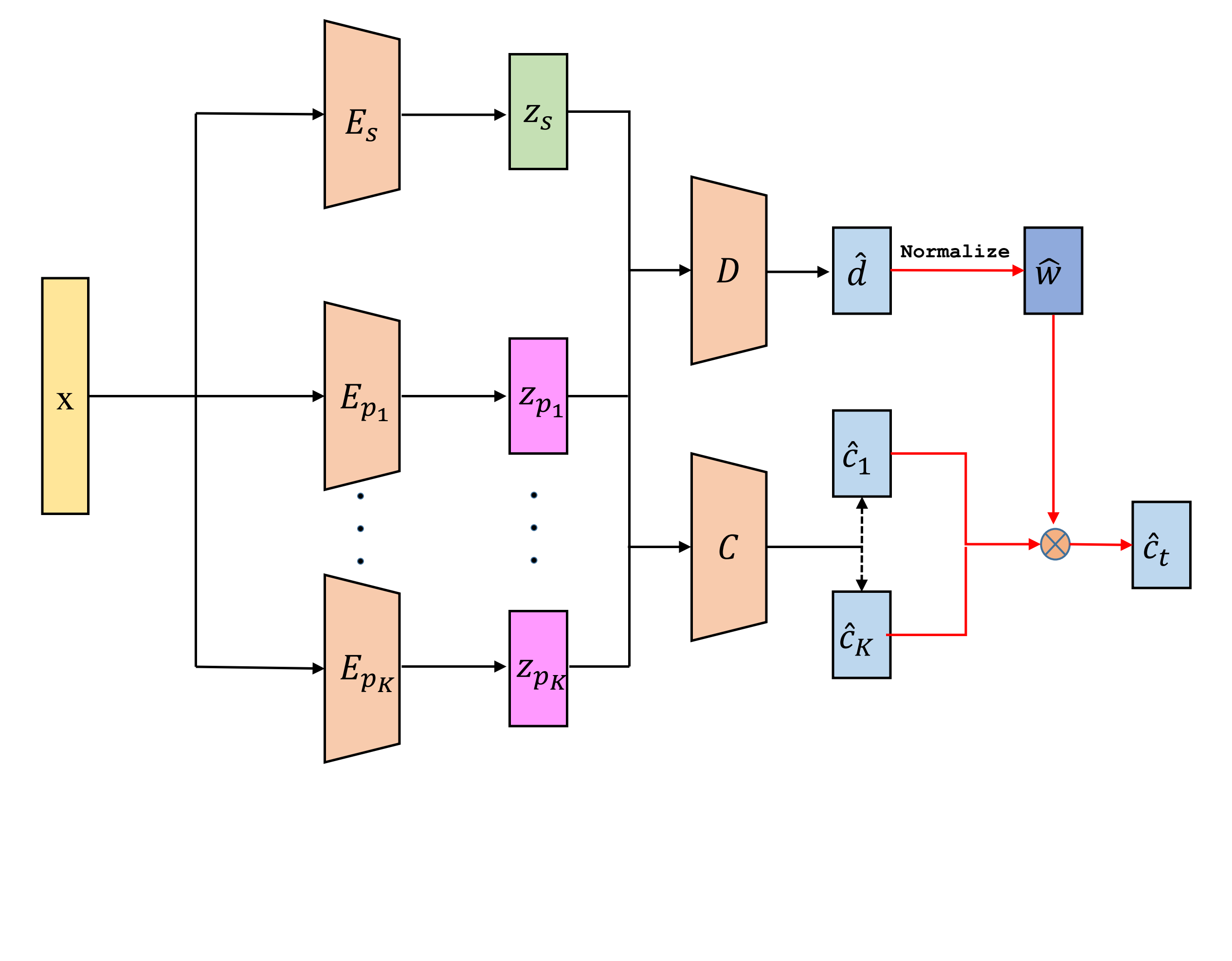}
  \vspace{-1.5cm}
  \caption{\first: The shared-wight extractor $E_s$ captures shared features $z_s$ for all domains. Each domain-specific extractor $\left \{ E_{p_j} \right \}_{j=1}^{K}$ captures private features $\left \{ z_{p_j} \right \}_{j=1}^{K}$ for each source domain. The classifier $D$ strives to discriminate which domain the instances coming from and forces $z_s$ domain-invariant and $\left \{ z_{p_j} \right \}_{j=1}^{K}$ domain-informative. 
  The $C$ estimates sentiment polarities $\left \{ \hat{c_{j}} \right \}_{j=1}^{K}$ from the views of different source domains as a traditional classifier. $\hat{d}$ is normalized to tell what confidence we can give to each $\left \{ \hat{c_{j}} \right \}_{j=1}^{K}$. Finally, the sentiment polarities of target domain are assembled by the weighted sum of $\left \{ \hat{c_{j}} \right \}_{j=1}^{K}$.
  }
  \label{framework}
\end{figure}
\begin{algorithm}[t]
\caption{Weighting Scheme based Unsupervised Domain Adaptation framework (\first)}
\label{alg1}
\textbf{Require}: labeled SOURCE corpus $\left \{\mathcal{S}\right \}$; unlabeled TARGET corpus $\left \{ \mathcal{T}\right \}$; all the DOMAIN labeled corpus $\left \{\mathcal{U}\right \}$; Hyperparameter $b\in\mathbb{N},\lambda>0, n_{critic}\in\mathbb{N}$
\begin{algorithmic}[1]
\State {Initialize parameters of $E_s, \left \{ E_{p_j} \right \}_{j=1}^{K}, D, C$}
\Repeat   
    \State    $\triangleright D$ iterations
    \For {$t = 1$\ \textbf{to} \ $n_{critic}$}
        \State $\ell_{\mathcal{D}} = 0$
        \ForAll {$d\in \mathcal{U}$}    
        \hfill {$\triangleright$ For all domains }
            \State {Sample a mini-batch $\boldsymbol{x} \triangleq (\boldsymbol{x}_{i},\boldsymbol{d}_{i})_{i=1}^{b} \sim \mathcal{U}_d$}
            
            \State{${{loss}}_{s} = \mathcal{L}_{D}(D(E_{s}(\boldsymbol{x}));d)$} 
            \State{$loss_{p_j} = 0$}
            \If{$d \notin \mathcal{T}_d$}
                \State{${{loss}}_{p_j}=\mathcal{L}_{\mathcal{D}}(D(E_{p_j}(\boldsymbol{x}));d)$}
            \EndIf
            \State {$\ell_{\mathcal{D}} \mathrel{+}= {{loss}}_{s}+{{loss}}_{p_j}$}
        \EndFor
        \State {Updating $D$ parameters to minimize  $\ell _{\mathcal{D}}$}
    \EndFor
    
    \State $\triangleright$Main  iterations
    \State $loss=0$
    \ForAll {$d\in \mathcal{S}$}       
    \hfill{$\triangleright$ For all $source$ domains}
        \State {Sample a mini-batch $\boldsymbol{x} \triangleq (\boldsymbol{x}_{i},\boldsymbol{y}_{i})_{i=1}^{b} \sim \mathcal{S}_d$}
        \State {$loss \mathrel{+}= \mathcal{L}_{C}(C( E_s(\boldsymbol{x}),E_{p_j}(\boldsymbol{x}));\boldsymbol{y})$}
    \EndFor
    
    \ForAll {$d\in \mathcal{U}$}      
    \hfill {$\triangleright$ For all domains}
        \State {Sample a mini-batch $\boldsymbol{x} \triangleq (\boldsymbol{x}_{i},\boldsymbol{d}_{i})_{i=1}^{b} \sim \mathcal{U}_d$}
        
        
        \State {$loss \mathrel{+}= -\lambda \cdot\mathcal{L}_{D}(D(E_{s}(\boldsymbol{x}));d)$}
    \EndFor
    
    \State {Updating $E_s, E_{p_j}, C$ parameters to minimize $loss$}
\Until {convergence}
        \end{algorithmic}
\end{algorithm}

In this section, we will present how to combine the $K$ source classifiers to derive a good target classifier.


Assume we have an unlabeled instance $x_i$ from target domain and need the decision value $f_i^{\mathcal{T}}=f^{\mathcal{T}}(x_i)$ of target classifier. To achieve this, we adopt a weighted combination of the $K$ source domain classifiers $f^j$ ($f_i^{j}=f^{j}(x_i)$) to approximate the target classifier. Specially, the approximated label $\left(\hat{y}_{i}\right)$  of the unlabeled target data $x_i$ is given by:

\begin{equation}
    \hat{y}_{i}=\sum_{j=1}^{K} w_i^{j} f_{i}^{j},
\end{equation}
where $w_i^{j}$ is a weight or instance-to-domain relation for the $i$-th target instance to $j$-th source domain. 

\first \  is modeled to train source classifiers simultaneously and integrate them to derive a good target classifier $f^{\mathcal{T}}$. Here, the two most important points are how to acquire a strong system which can benefit every $f^{j}$ and obtain $w_i^j$ according to the relations between target instances and source domains. For the first point, we employ an effective multi-task architecture named shared-private model~\cite{bousmalis2016domain,liu2017adversarial,gholami2018unsupervised}; For the second point, we introduce a weighting scheme to 
estimate $w_i^{j}$ by utilizing the discriminator as a probability distribution estimator. As Figure~\ref{framework} illustrates, this framework includes $K$ domain-specific extractors $\left \{ E_{p_j} \right \}_{j=1}^{K}$, a shared feature extractor $E_s$, a classifier $C$, and a discriminator $D$. 
In detail, we exploit $E_s$ to distill domain-invariant features and $\left \{ E_{p_j} \right \}_{j=1}^{K}$ to extract domain-informative features for each domain through a constraint defined by an elaborately designed discriminator $D$.
At final, we concatenate the shared and private features to obtain the sentiment polarities through the classifier $C$. 

In the following, we will describe the input, objective, motivation and the training process for each module respectively. 
\subsubsection{The discriminator $D$}
In general, we assume the existence of a shared feature space between domains where the distribution divergence is small and reducing domain divergence through the adversarial training upon the deep learning framework can further exploit domain invariant features. We utilize a discriminator $D$ to implement adversarial training and obtain the corresponding loss. The objective of discriminator $D$ is:
\begin{equation}\label{Dis}
    \begin{aligned}
    \mathcal{L}_{D} = &-\frac{1}{N} \sum_{i=1}^{N} \boldsymbol{d}_{i}^{\top} \ln D\left(E_{s}\left(\boldsymbol{x}_{i}\right)\right)\\
    &-\frac{1}{N_{s}} \sum_{j=1}^{K} \sum_{i=1}^{\left|S_{j}\right|} \boldsymbol{d}_{i}^{\top} \ln D\left(E_{p_{j}}\left(\boldsymbol{x}_{i}\right)\right),
    \end{aligned}
\end{equation}

where 
$\boldsymbol{d}_{i}$ is the domain label and $D(\cdot)$ can be considered as a classifier, which will output the label probabilities of domain prediction. After convergence of the adversarial training, $D$ can discriminate which domain the features coming from. The two terms signify that $D$ needs to discriminate the two parts of features separately.

It must be stressed that $D$ has three key roles. The first is to help acquire more pure shared features. The second is to help fulfill the separation of shared and private features through its discriminated ability. The last but not least is to measure how likely the target instance is coming from the corresponding source domains
, which implicitly introduces the data-dependent priors of source domains.

\subsubsection{The classifier $C$}
$C$ is an usual classifier and used to classify sentiment polarities. Its objective is: 

\begin{equation}\label{cla}
    \mathcal{L}_{C} =-\frac{1}{N_s} \sum_{j=1}^{K}\sum_{i=1}^{\left | S_j \right |} \boldsymbol{y}_{i}^{T} \ln C\left(E_{s}\left(\boldsymbol{x}_{i}\right), E_{p_j}\left(\boldsymbol{x}_{i}\right)\right),
\end{equation}

where 
$\boldsymbol{y}_{i}$ is the class label and $C(\cdot)$ is a function of classical classifier. This objective means the concatenated features will be sent to $C$ to calculate the final sentiment label.
It is noticeable, however, that if we set all the private features to zeros or abandon them, such as previous methods did~\cite{liu2017adversarial,chen2018multinomial,zhao2018multiple}, will lead to knowledge loss and the system performance will decrease.
\subsubsection{The private feature extractor $\left \{ E_{p_j} \right \}_{j=1}^{K}$}
As discussed by~\citeauthor{salzmann2010factorized,bousmalis2016domain,liu2017adversarial}, the independence of private features and shared features is beneficial for system performance. If we just utilize the discriminator to constrain the shared feature extractor, the domain-related knowledge will appear both in shared space and private space. To overcome this, \citeauthor{bousmalis2016domain,liu2017adversarial} adopt the orthogonality constraints, which penalize redundant latent representations and encourages the shared and private extractors to encode different aspects of the inputs. Here, we deeply exploit $D$ to make a constraint, reserve the private characteristic of each domain in their private space
and encourage moving the common information from each domain to their shared space. The objective of each private extractor $E_{p_j}$ is:
\begin{equation}
\label{e3}
\begin{aligned}
    \mathcal{L}_{P}=&-\frac{1}{N_s} \sum_{j=1}^{K}\sum_{i=1}^{\left | S_j \right |} \boldsymbol{y}_{i}^{T} \ln C\left(E_{s}\left(\boldsymbol{x}_{i}\right), E_{p_j}\left(\boldsymbol{x}_{i}\right)\right)\\
    &-\frac{1}{N_s} \sum_{j=1}^{K}\sum_{i=1}^{\left | S_j \right |} \boldsymbol{d}_{i}^{\top} \ln D\left(E_{p_j}\left(\boldsymbol{x}_{i}\right)\right).
\end{aligned}
\end{equation}
The private feature extractors have two aspects of roles, one is to be concatenated with shared features to contribute to the polarity classification, and the other is trying its best to retain more domain-related information for $D$ to discriminate the domain labels, which are depicted by the first term and the second term separately in Equation~\ref{e3}. 
\subsubsection{The shared feature extractor $E_s$}
Through the adversarial training, the distribution divergence between different domains in the shared feature space will become a minimum and the features extracted by $E_s$ will be domain-invariant and can be shared across all the domains. The objective is:
\begin{equation}
    \begin{aligned}
        \mathcal{L}_{S}=&-\frac{1}{N_s} \sum_{j=1}^{K}\sum_{i=1}^{\left | S_j \right |} \boldsymbol{y}_{i}^{T} \ln C\left(E_{s}\left(\boldsymbol{x}_{i}\right), E_{p_j}\left(\boldsymbol{x}_{i}\right)\right)\\
        &+\frac{1}{N} \sum_{i=1}^{N} \boldsymbol{d}_{i}^{\top} \ln D\left(E_{s}\left(\boldsymbol{x}_{i}\right)\right).
    \end{aligned}
\end{equation}
The first term denotes the contribution to the sentiment classification when concatenated with the private features. The second term is the adversarial loss, which will encourage $E_s$ to strengthen its feature extraction ability to confuse $D$. After reaching the Nash Equilibrium Point, features extracted by $E_s$ can not be discriminated by $D$. 
\subsubsection{The training process}
In this subsection, we will show how to integrate all modules and train them. Overall, we follow the training tricks adopted by~\citeauthor{arjovsky2017wasserstein}~(\citeyear{arjovsky2017wasserstein}) and update parameters in two stages. The first stage is to enhance the discriminant ability of $D$ by minimizing the adversarial loss.
At the second stage, we backward the classification error and update the parameters of $E_s$ and $E_{p_j}$. The whole training process is demonstrated in Algorithm~\ref{alg1}. 

After the convergence of above training, we can directly obtain the labels estimated by the pre-trained source classifiers and do not need any further training. At inference time, $D$ will be utilized as a  probability  distribution estimator and measures the instance-to-domain relations (i.e. $\hat{\boldsymbol{w}}_{\boldsymbol{j}}$). The final target labels can be calculated by:
\begin{equation}\label{ct}
    \hat{\boldsymbol{c}}_{t}=\sum_{j=1}^{K} \hat{\boldsymbol{c}}_{\boldsymbol{j}} \cdot \operatorname{Normalize}\left(\hat{\boldsymbol{w}}_{\boldsymbol{j}}\right).
\end{equation}
\subsection{The overview of \second}
In this section, we will introduce the overview and training process for our second framework. 

As described before, our first framework can annotate target examples with pseudo labels directly by the source classifiers. However, the performance of this framework will not improve with access to more confident pseudo labels or real labeled data. Motivated by the multi-view training methods~\cite{blum1998combining,zhou2005tri,saito2017asymmetric}, we propose the second framework to train a target-specific extractor by further exploiting pseudo labels, whose performance can progress accompanying with the increase of labeled data. Nevertheless, it is worth noting that the pseudo labels can be wrong and will decrease the performance quickly if not well handled. To deal with this problem, we adopt the same architecture as the first framework but different training mechanism, which gradually adds the pseudo labels in the training process.

\subsubsection{The training process}
\begin{algorithm}[t]
\caption{Two-stage Training based Unsupervised Domain Adaptation (\second) }
\label{alg2}
\textbf{Require}: unlabeled TARGET corpus $\left \{ \mathcal{T}\right \}$; pre-trained model $E_s, E_{p_j}, D, C$; hyperparameter $b\in\mathbb{N}$
\begin{algorithmic}[1] 
\State{Load parameters of $E_s, \left \{ E_{p_j} \right \}_{j=1}^{K}, D, C$} 
\State{Initialize parameters of $E_t, \eta =0.02, \Delta=0.98, N=10, \mathcal{T}_l = \varnothing, \mathcal{L} = \varnothing$}
\Repeat
\State{Sample a mini-batch $\boldsymbol{x}_t \triangleq (\boldsymbol{x}_{i},\boldsymbol{d}_{i})_{i=1}^{b} \sim \mathcal{T}$}
\State{$\boldsymbol{\hat{y}}_{i}^{\mathcal{S}}= labeling(E_{s}\left(\boldsymbol{x}_{t}\right), E_{p_{j}}\left(\boldsymbol{x}_{t}\right))$}
\State{$\boldsymbol{\hat{y}}_{i}^{\mathcal{T}}=C(E_s(\boldsymbol{x}_t),E_{t}(\boldsymbol{x}_t))$}
\State{$\boldsymbol{\hat{y}}_{i}=\boldsymbol{\hat{y}}_{i}^{\mathcal{S}}\cap \boldsymbol{\hat{y}}_{i}^{\mathcal{T}}$  \ \ \ }
\State{$\mathcal{L}\triangleq\left(\boldsymbol{x}_{i}, \hat{\boldsymbol{y}}_{i}\right)$}
\For {$j = 1$\ \textbf{to} \ $iter$}        \hfill {$\triangleright$ Train $E_t$ with $\mathcal{L}$}
\State{Sample a mini-batch $\boldsymbol{\hat{x}}_{t} \triangleq (\boldsymbol{\hat{x}}_{i}, \boldsymbol{\hat{y}}_{i})_{i=1}^{b} \sim \mathcal{L}$}
\State{$loss \mathrel{=} \mathcal{L}_{C}(C(E_s(\boldsymbol{\hat{x}}_{t}),E_{t}(\boldsymbol{\hat{x}}_{t}));\boldsymbol{\hat{y}})$}
\State{Update $E_t$ by using $\nabla loss$}
\EndFor
\State{$\mathcal{T}_l=\mathcal{T}_l\cup \mathcal{L}$}
\State{$\mathcal{T} = \mathcal{T} \ \backslash \ \mathcal{L}$}
\State{$\Delta =\Delta -\eta $}
\Until{$\left | \tau^{-1} \right | + \left | \tau^{-2}  \right | \leq N \ or \ \Delta \leq 0.5$}
\Repeat
\For {$j = 1$\ \textbf{to} \ $iter$}        \hfill {$\triangleright$ Train $E_t$ with $\mathcal{T}_l$}
\State{Sample a mini-batch $\boldsymbol{\hat{x}}_{t} \triangleq (\boldsymbol{\hat{x}}_{i}, \boldsymbol{\hat{y}}_{i})_{i=1}^{b} \sim \mathcal{T}_l$}
\State{$loss \mathrel{=} \mathcal{L}_{C}(C(E_s(\boldsymbol{\hat{x}}_{t}),E_{t}(\boldsymbol{\hat{x}}_{t}));\boldsymbol{\hat{y}})$}
\State{Update $E_t$ by using $\nabla loss$}
\EndFor
\Until {convergence}

\end{algorithmic}
\end{algorithm}
 To acquire more confident pseudo labels gradually, at the beginning we just trust the labels whose confidence is greater than a dynamic threshold $\Delta$ (e.g. 0.98), which is changing while training. In the training process, we will lessen the confidence to some degree, such as every time decrease a constant $\eta =0.02$. When the number of the pseudo-label set, which we denote as $ \mathcal{T}_l $, does not increase obviously, we will terminate the training process at the optimal confidence. For training, we use $\tau^{-1}$ and $\tau^{-2}$ to represent the pseudo-label set generated by two successive iterations and control the terminal condition. As long as the number of pseudo labels becomes less than a certain number, which is represented by $N$(e.g. 10), then the iteration will terminate. Besides, we denote $iter$ as the iteration of training and the function $labeling$ means the method of labeling by source classifiers. We assign pseudo-labels for target samples when the weighted predictions produced by all source classifiers and the predicted results of the target classifier are both bigger than $\Delta$ (i.e. $\hat{\boldsymbol{c}}_{t}>\Delta$, where $\hat{\boldsymbol{c}}_{t}$ is calculated as Equation~\ref{ct}). Then we use these samples to train the target feature extractor. With the acquirement of all pseudo labels, we will finetune the target feature extractor.
 More detailed training information is presented in Algorithm~\ref{alg2}.


\section{Experiments}
In this section, we empirically evaluate the performance of the proposed frameworks.
\subsection{Experimental Settings}
\subsubsection{Amazon review dataset~\cite{blitzer2007biographies}}
We conduct the experiments on the Amazon reviews dataset, which has been widely used for cross-domain sentiment classification. As described by \citeauthor{chen2018multinomial}, the pre-processed features loss the order information, which prohibits the usage of strong feature extractor (e.g. RNN or CNN). For fair comparison, we also adopt a MLP as our feature extractor and
represent each review as a 5000d feature vector which are the most frequent features.

The Amazon dataset contains 2000 samples for each of the four domains: $book$, $DVD$, $electronics$, and $kitchen$, with binary labels (positive, negative), and more details are included in Table~\ref{amazon}. 
\begin{table}
\centering
\begin{tabular}{l|cccc}
\hline
\textbf{Domain} & \textbf{\# Labeled} & \textbf{\# Unlabeled} & \textbf{\% Negative} \\
\hline
book & 2000  & 4465 & 49.29     \\
DVD & 2000 & 3586 & 49.61      \\
electronics & 2000 & 5681 & 49.71   \\
kitchen & 2000 & 5945 & 50.31   \\
\hline
\end{tabular}
\caption{ Statistics of the Amazon reviews dataset including the number of labeled and unlabeled reviews for each domain as well as the proportion of negative samples in the unlabeled data.}
\label{amazon}
\end{table}

\subsubsection{FDU-MTL~\cite{liu2017adversarial}}
As described before, the Amazon reviews dataset has many limitations, especially the reviews are already tokenized and converted
to a bag of features and lack of the order information. To further validate our frameworks, we turn to another dataset with raw review texts, which is in line with the actual application scenario. With the raw review texts, we can process them from the very start and many strong feature extractor architectures can be adopted (e.g. LSTM). This dataset has 16 different domains of reviews: $Books$, $electronics$, $DVD$, $kitchen$, $apparel$, $camera$, $health$, $music$, $toys$, $video$, $baby$, $magazine$, $software$, and $sports$, in addition to two movies review domains from the IMDb and the MR dataset. The amount of training and unlabeled data of each domain vary across domains but are roughly 1400 and 2000, respectively. In addition, each domain has a development set of 200 samples and a test set of 400 samples. 
\subsection{Implementation Details}
For both datasets, we take turns selecting one domain as the target domain and the remained domains as the source domains. Although we consider just one domain as the target domain, our framework can be easily extended to handle multiple target domains.
\subsubsection{Details on Amazon}
Firstly, we use the data (both labeled and unlabeled) in the source domains and the data (only unlabeled) in the target domain to complete the training of the main process in Algorithm~\ref{alg1}. And then, we randomly select 2000 samples from the target domain to be annotated for both frameworks, while for \second \ different training mechanism is adopted (i.e. Algorithm~\ref{alg2}) to train the target-specific private feature extractor. For both frameworks, the remaining samples in the target domain are used as the validation set and test set, and the number of samples in the validation set is the same as (Chen  et  al.  2018). The private feature extractor is optimized with the Adam over shuffled mini-batches. And we set the batch size 8 and the learning rate 0.0001 for the sentiment classifier and the domain classifier. Besides, we perform early stopping on the validation set during the training process.
\subsubsection{Details on FDU-MTL}
For this dataset, we randomly select about 1400 samples out from target domain as alternative pseudo labels. And the number of samples in the validation set and test set is consistent with that in the (Chen et al. 2018). Other implementation details and the rest of the parameter settings are the same as those for the Amazon dataset.
\subsection{Performance Comparison}
For all the experiments, we use classification accuracy to measure the performance of our frameworks.
\subsubsection{Amazon}
When comparing with single-source domain adaptation methods (i.e. mSDA, DANN), the training data in the multiple source domains are
combined and viewed as a single domain. All baseline methods in the comparison include:
\begin{itemize}
    \item \textbf{mSDA~\cite{chen2012marginalized}}: it employs marginalized stacked denoising autoencoders to learn new representations for domain adaptation. Specially, this method does not require stochastic gradient descent or other optimization algorithms to learn parameters.
    \item \textbf{DANN~\cite{ganin2016domain}}: it introduces a new representation learning approach for domain adaptation, which is based on the adversarial training. 
    \item \textbf{MDAN(H-MAX), MDAN(S-MAX)~\cite{zhao2017multiple}}: they are two adversarial neural models. One directly optimizes the proposed new generalization bound (i.e. H-MAX) and another is a smoothed approximation of the first one (i.e. S-MAX), leading to a more data-efficient and task-adaptive model. These two methods are in multi-source-to-one-target setting. 
\end{itemize}
\begin{table}
\centering
\setlength{\tabcolsep}{0.7mm}{
\begin{tabular}{l|cccc|c}
\hline
Target Domain  & Book & DVD & Elec. & Kit  & Avg \\
\hline
MLP & 76.55  & 75.88 & 84.60 & 85.45 & 80.46     \\
mSDA & 76.98 & 78.61 & 81.98 & 84.26 & 80.46      \\
DANN & 77.89 & 78.86 & 84.91 & 86.39 & 82.01 \\
MDAN(H-MAX) & 78.45 & 77.97 & 84.83 & 85.80 & 81.76 \\
MDAN(S-MAX) & 78.63 & 80.65 & \textbf{85.34} & 86.26 & 82.72 \\
MAN-L2 & 78.45 & 81.57 & 83.37 & 85.57 & 82.24 \\
MAN-NLL & 77.78 & 82.74 & 83.75 & 86.41 & 82.67     \\
\hline
\first & 79.39 & 80.14 & 83.81 & 87.66 & 82.75 \\
          &\scriptsize (+0.76) &\scriptsize (-2.60) &\scriptsize (-1.53) & \scriptsize (+1.25) & \scriptsize (+0.03) \\
\second & \textbf{79.92} & \textbf{83.86} & 85.11 & \textbf{87.68} & \textbf{84.14}     \\
        &\scriptsize (+1.29) &\scriptsize (+1.12) &\scriptsize (-0.23) & \scriptsize (+1.27) & \scriptsize (+1.42) \\
\hline
\end{tabular}}
\caption{Results on Amazon review dataset. The highest domain performance is shown in bold. The numbers in brackets represent the improvements relative to the best performance in the above baselines. Except for our model, the rest is taken from ~\cite{chen2018multinomial}.}
\label{tab2}
\end{table}
Table~\ref{tab2} reports the classification accuracy of different methods on the Amazon reviews dataset. It shows that our second framework outperforms the state of art by 1.42 percentage. But for the first framework, we just obtain a comparable average result. For this, we argue the reason is that we can not adopt strong source extractors, the integrated results will also be weak.  
It is worthwhile noting that all the baselines are just for one target domain, but our framework can deal with multiple target domains.
\subsubsection{FDU-MTL}
\begin{table*}
\centering 
\setlength{\tabcolsep}{0.7mm}{
\scalebox{1.0}{
\begin{tabular}{l|cccccccccccccccc|c}
\hline
Target  & books & elec.  &  dvd  &  kitchen  &  apparel & camera  &  health  & music &  toys &  video  &  baby&magaz. &sofw. &sports & IMDb &MR & Avg \\
\hline
$ASP^1$  & 83.2  & 82.2  & 85.5  & 83.7 & 87.5  & 88.2  & 87.7  & 82.5  & 87.0  & 85.2 & 86.5 &91.2 &85.5 &86.7 &87.5 &75.2 &85.3     \\
$ASP^2$ & 83.7  & 83.2 & 85.7  & 85.0  & 86.2 & 89.7  & 86.5  & 81.7  & 88.2  & 85.2 & 88.0 &90.5 & 88.2 &86.5 &86.7 &76.5 &85.7      \\
$MAN$ & 84.5  & 86.5  & 86.3  & 87.3  & 86.0 & 83.8  & 88.5  & 84.0  & 87.3  &87.0 & 87.0 &86.8 &86.3 &88.3 &84.3 &73.3 &85.4 \\
$Meta$ &\textbf{86.3}  & 86.0 &86.5 &86.3 &86.0 & 87.0 &\textbf{88.7} &\textbf{85.7} &85.3 & 85.5 & 86.0 &\textbf{90.3} &86.5 &85.7 &87.3 & \textbf{75.5} &85.9\\
\hline
\first & 84.6 & 87.9 &\textbf{88.9} &88.7 &\textbf{90.1} &\textbf{90.2} &87.8 &84.3 &\textbf{89.1} &\textbf{88.7} & 87.4 & 86.6 & 88.4 & 88.2 &\textbf{88.7} & 74.6 &87.1 \\
          & \scriptsize (-1.7) & \scriptsize (+1.9) &\scriptsize (+2.4) &\scriptsize (+2.4) &\scriptsize (+4.1) & \scriptsize (+3.2) & \scriptsize (-0.9) &\scriptsize (-1.4) &\scriptsize (+3.8) & \scriptsize (+3.2) & \scriptsize (+1.4) & \scriptsize (-3.7) & \scriptsize (+1.9) & \scriptsize (+2.5) &\scriptsize (+1.4) & \scriptsize (-0.9) &\scriptsize (+1.2) \\
\second& \textbf{86.3} & \textbf{89.5} & 88.5 &\textbf{89.3} &89.5 &89.1 & 88.5 &84.1 &89.0 & 87.5 &\textbf{89.0} & 88.8 & \textbf{88.8} & \textbf{88.3} &88.3 & 73.3 &\textbf{87.4}     \\
        & \scriptsize (+0.0) & \scriptsize (+3.5) &\scriptsize (+2.0) &\scriptsize (+3.0) &\scriptsize (+3.5) & \scriptsize (+2.1) & \scriptsize (-0.2) &\scriptsize (-1.6) &\scriptsize (+3.7) & \scriptsize (+2.0) & \scriptsize (+3.0) & \scriptsize (-1.5) & \scriptsize (+2.3) & \scriptsize (+2.6) &\scriptsize (+1.0) & \scriptsize (-2.2) &\scriptsize (+1.5) \\
\hline
\end{tabular}}}
\caption{Results on FDU-MTL dataset. The highest performance in each domain is highlighted. The numbers in brackets represent the improvements relative to the $Meta$ baseline. Results of 'MAN' are obtained by setting private features to zeros and other results are taken from their original papers.}
\label{tab3}
\end{table*}

The baseline methods in the comparison include:
\begin{itemize}
    \item \textbf{ASP-MTL-SC, ASP-MTL-BC~\cite{liu2017adversarial}}: these two models are single channel model  and  bi-channel  model  of adversarial multi-task learning.
    \item \textbf{MAN~\cite{chen2018multinomial}}: this model is a multinomial adversarial networks for multi-domain text classification, which is extended from \cite{liu2017adversarial}
    \item \textbf{Meta-MTL~\cite{chen2018meta}}: it is a meta-network, which can capture the meta-knowledge of semantic composition and generate the parameters of the task specific semantic composition models.
\end{itemize}

As shown in Table~\ref{tab3}, the performance of our frameworks performs better in 12 of 16 domains when comparing with all other algorithms. In average accuracy, we obtain 1.2 and 1.5 percent improvement for our two introduced frameworks separately.

\subsection{Discussion}
In this section, we will show the effectiveness of the feature separating ability and the validation of the weighting scheme in our frameworks.
\begin{figure}[htbp]
  \centering
  \subfigure{\label{dis1}\includegraphics[scale=0.33]{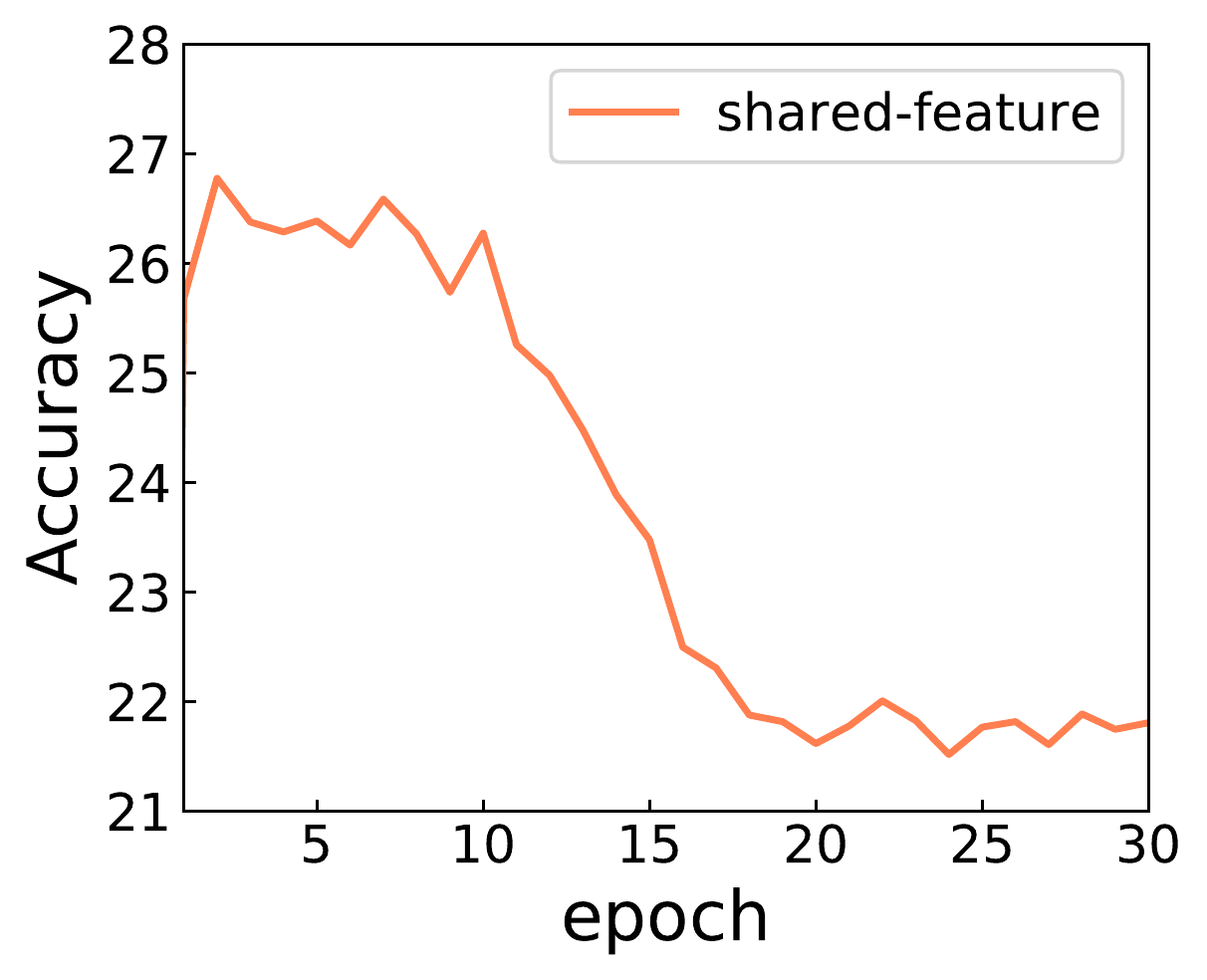}}
  \subfigure{\label{dis2}\includegraphics[scale=0.33]{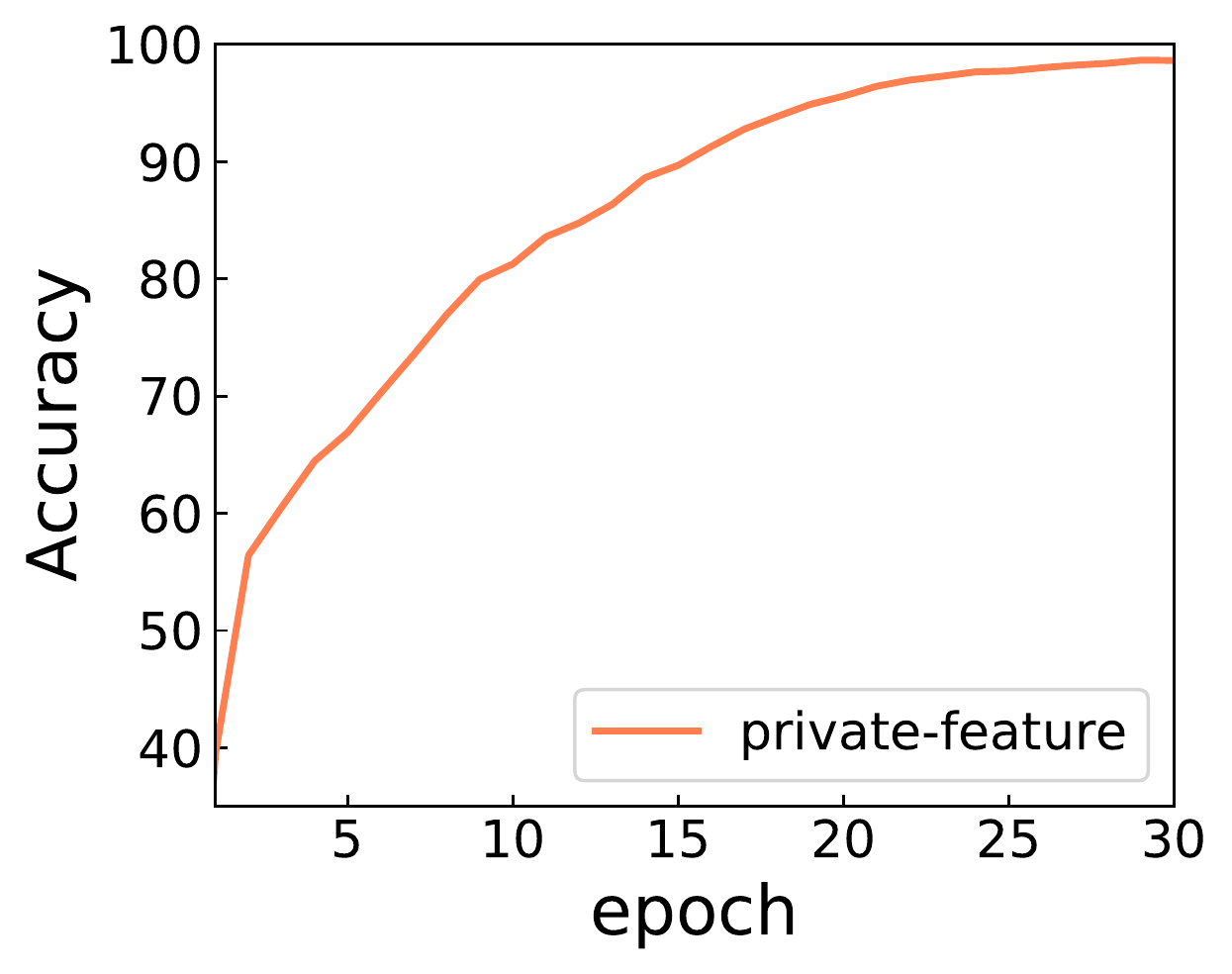}}
  \caption{The accuracy of $D$ for private features and shared features on Amazon when training under \first.}
\label{dis}
\end{figure}

\subsubsection{The effectiveness of separating ability}
In this subsection, we will show the accuracy of D in the training process for experiments on Amazon, which will reveal the discriminating ability of $D$ and the feature extraction ability of $E_s$ and $E_{p_j}$. 
As Figure~\ref{dis} depicts, the accuracy of $D$ for shared feature roughly converges at $1/K$ and for private features stabilises at around 90 percentage. The results tell that in the training process the discriminated ability of $D$ will become stronger and stronger, which will force $E_s$ to extract more pure shared features and $E_{p_j}$ to retain most of the domain-related information.
\subsubsection{The validation of weighting scheme}
Table~\ref{tab3} reveals the effectiveness of the weighting scheme in our frameworks in total, to further demonstrate its effectiveness, we will pick out some representative examples (i.e. some sentences) to visualize the weights assigned to each source classifier.

We use Figure 3 to show the weight assigned to each source classifier from two sentences picking from the 'toy' and 'dvd' domain. As demonstrated in ~\cite{ben2007analysis,blitzer2007biographies}, a metric named $\mathcal{A}$-distance can be utilized to well measure the relation between two domains in an unsupervised way. We measure the distances between 'toy' and 'dvd' with other domains, the ranking from close to not related are \textit{baby, sports\_outdoors, health\_personal\_care} and \textit{video, books, imdb} respectively. And our weights tell the rankings are \textit{baby, sports\_outdoors, music} and \textit{video, books, magazines}. The overall rankings are coincident with the $\mathcal{A}$-distance.

\begin{figure}[htbp]
  \centering
  \subfigure{\label{toy}\includegraphics[scale=0.26]{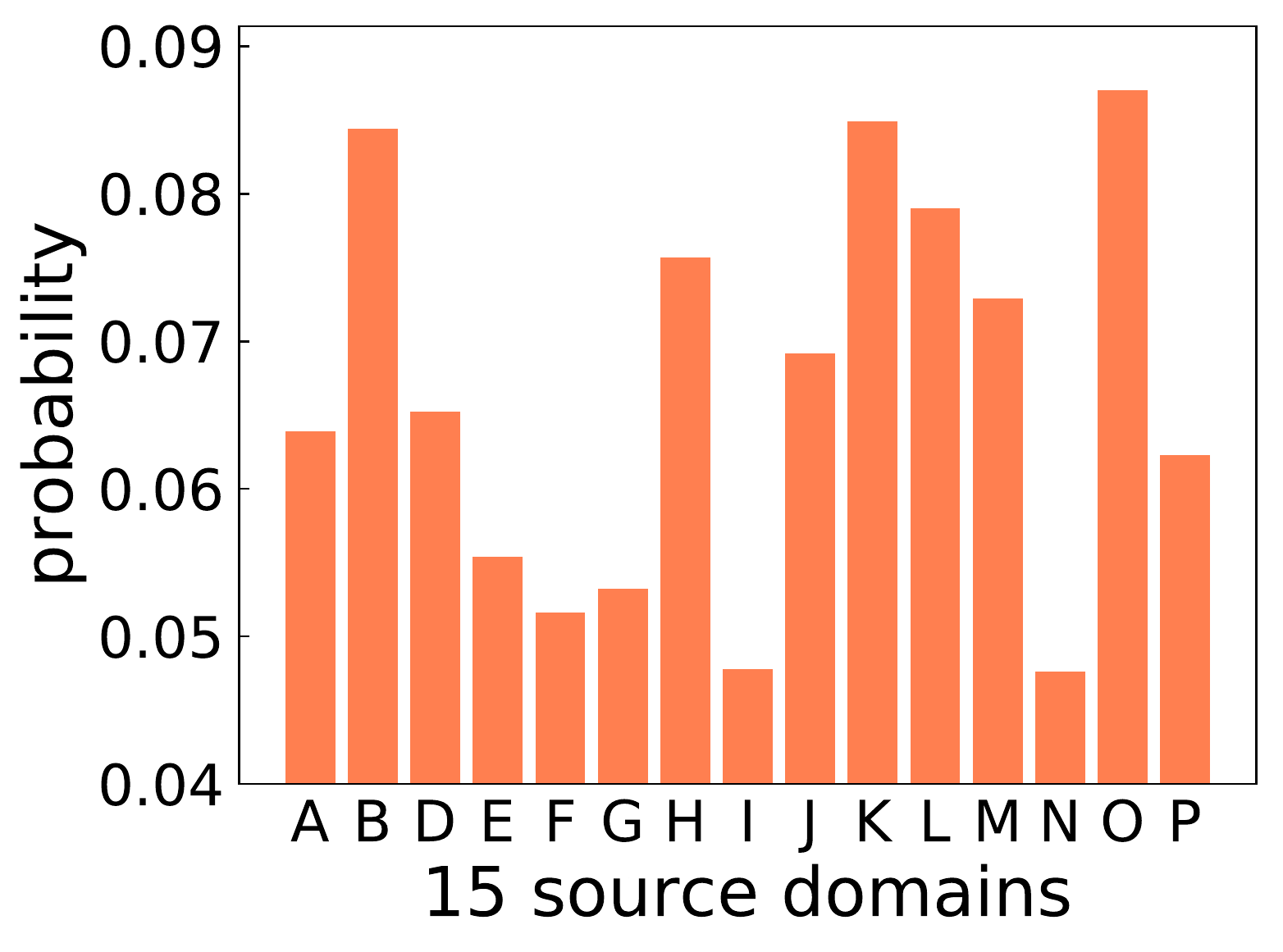}}
  \subfigure{\label{dvd}\includegraphics[scale=0.26]{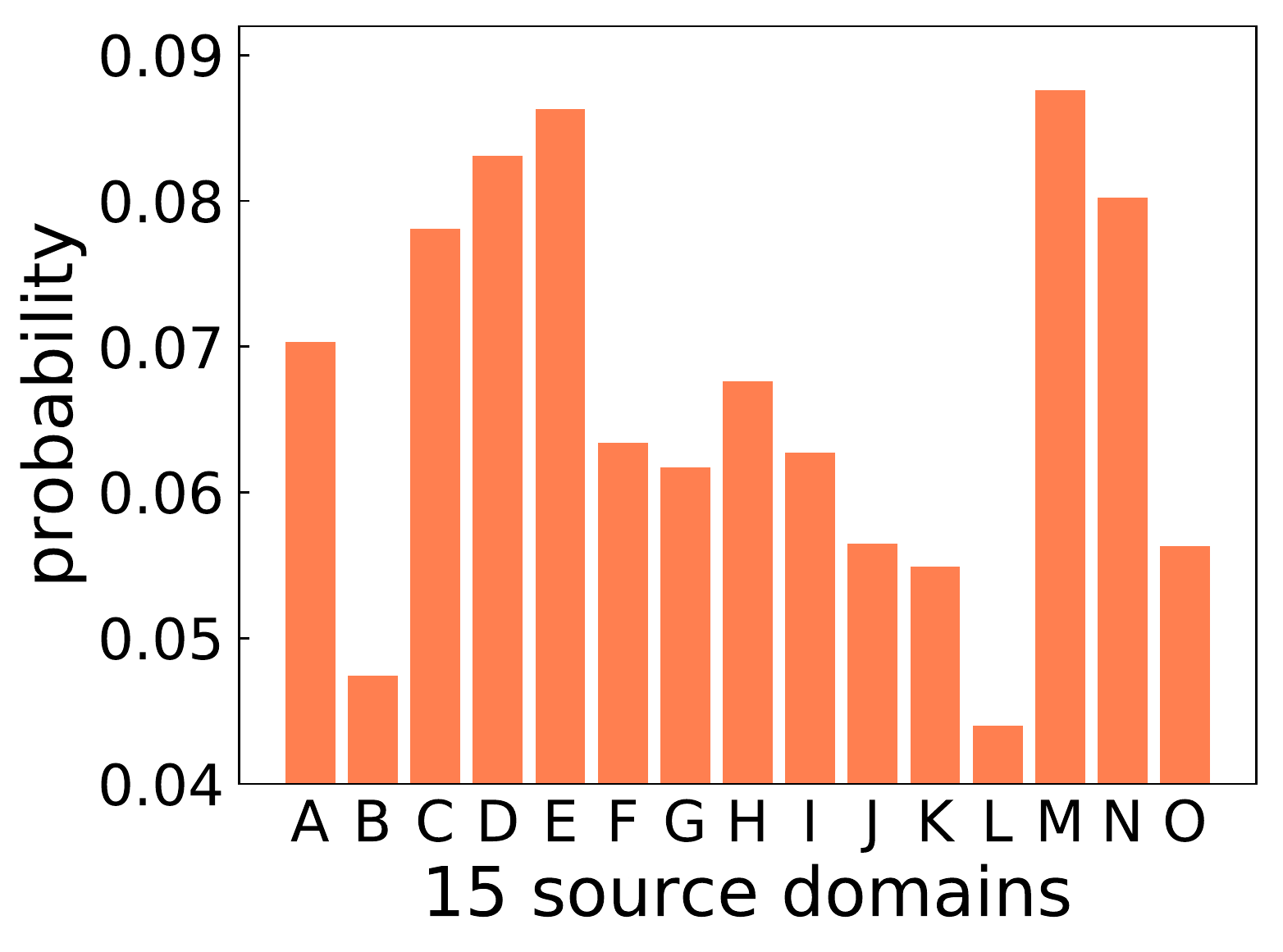}}
  \caption{The weight assigned to each source classifier for sentiment classification from two sentences. The first sentence is  '\textit{i love this toy ! i 'm 12 years old and i still love to play barbie 's ! i would reccomend this toy to any mother who has a little girl or boy that loves to play barbie's}', which is derived from the 'toy' domain. The second sentence is '\textit{i must be missing something , i bought this movie ... when i watched it i fell asleep , its boring as hell trust me im a true horror fan and a gore-fiend , ... , the only scary part was the end with this witch , is everyone high but me ? this movie sucked ! boring as hell..}', which is derived from the 'dvd' domain. The characters 'A-P' denote different domains from 'MR' to 'dvd'.}
\label{example}
\end{figure}


\section{Conclusion}
In this paper, we introduce two frameworks to do SA in MS-UDA setting. The introduced frameworks adopt a novel weighting scheme for annotating pseudo labels and an effective training mechanism based on the pseudo labels to obtain a target-specific extractor. More importantly, the role of the discriminator as a probability distribution estimator is a key feature of our frameworks, which can be further studied in the future in other application scenarios. Experimental results on two SA datasets demonstrate the promising performance of our frameworks. The proposed frameworks could be conveniently adapted to other text classification tasks, or extended to multiple target domains setting.

\section{Acknowledgements}
This work was partially supported by NSF China (Nos.61572111 and G05QNQR004). We thank Dr. Pengfei Liu and Dr. Zirui Wang for an insightful discussion on the design of our frameworks.

\begin{thebibliography}{}

\bibitem[\protect\citeauthoryear{Arjovsky, Chintala, and
  Bottou}{2017}]{arjovsky2017wasserstein}
Arjovsky, M.; Chintala, S.; and Bottou, L.
\newblock 2017.
\newblock Wasserstein gan.
\newblock {\em arXiv preprint arXiv:1701.07875}.

\bibitem[\protect\citeauthoryear{Ben-David \bgroup et al\mbox.\egroup
  }{2007}]{ben2007analysis}
Ben-David, S.; Blitzer, J.; Crammer, K.; and Pereira, F.
\newblock 2007.
\newblock Analysis of representations for domain adaptation.
\newblock In {\em Advances in neural information processing systems},
  137--144.

\bibitem[\protect\citeauthoryear{Blitzer, Dredze, and
  Pereira}{2007}]{blitzer2007biographies}
Blitzer, J.; Dredze, M.; and Pereira, F.
\newblock 2007.
\newblock Biographies, bollywood, boom-boxes and blenders: Domain adaptation
  for sentiment classification.
\newblock In {\em Proceedings of the 45th annual meeting of the association of
  computational linguistics},  440--447.

\bibitem[\protect\citeauthoryear{Blum and Mitchell}{1998}]{blum1998combining}
Blum, A., and Mitchell, T.
\newblock 1998.
\newblock Combining labeled and unlabeled data with co-training.
\newblock In {\em Proceedings of the eleventh annual conference on
  Computational learning theory},  92--100.
\newblock Citeseer.

\bibitem[\protect\citeauthoryear{Bousmalis \bgroup et al\mbox.\egroup
  }{2016}]{bousmalis2016domain}
Bousmalis, K.; Trigeorgis, G.; Silberman, N.; Krishnan, D.; and Erhan, D.
\newblock 2016.
\newblock Domain separation networks.
\newblock In {\em Advances in neural information processing systems},
  343--351.

\bibitem[\protect\citeauthoryear{Chattopadhyay \bgroup et al\mbox.\egroup
  }{2012}]{chattopadhyay2012multisource}
Chattopadhyay, R.; Sun, Q.; Fan, W.; Davidson, I.; Panchanathan, S.; and Ye, J.
\newblock 2012.
\newblock Multisource domain adaptation and its application to early detection
  of fatigue.
\newblock {\em ACM Transactions on Knowledge Discovery from Data (TKDD)}
  6(4):18.

\bibitem[\protect\citeauthoryear{Chen and Cardie}{2018}]{chen2018multinomial}
Chen, X., and Cardie, C.
\newblock 2018.
\newblock Multinomial adversarial networks for multi-domain text
  classification.
\newblock {\em arXiv preprint arXiv:1802.05694}.

\bibitem[\protect\citeauthoryear{Chen \bgroup et al\mbox.\egroup
  }{2012}]{chen2012marginalized}
Chen, M.; Xu, Z.; Weinberger, K.; and Sha, F.
\newblock 2012.
\newblock Marginalized denoising autoencoders for domain adaptation.
\newblock {\em arXiv preprint arXiv:1206.4683}.

\bibitem[\protect\citeauthoryear{Chen \bgroup et al\mbox.\egroup
  }{2016}]{chen2016neural}
Chen, H.; Sun, M.; Tu, C.; Lin, Y.; and Liu, Z.
\newblock 2016.
\newblock Neural sentiment classification with user and product attention.
\newblock In {\em Proceedings of the 2016 Conference on Empirical Methods in
  Natural Language Processing},  1650--1659.

\bibitem[\protect\citeauthoryear{Chen \bgroup et al\mbox.\egroup
  }{2018}]{chen2018meta}
Chen, J.; Qiu, X.; Liu, P.; and Huang, X.
\newblock 2018.
\newblock Meta multi-task learning for sequence modeling.
\newblock In {\em Thirty-Second AAAI Conference on Artificial Intelligence}.

\bibitem[\protect\citeauthoryear{Duan \bgroup et al\mbox.\egroup
  }{2009}]{duan2009domain}
Duan, L.; Tsang, I.~W.; Xu, D.; and Chua, T.-S.
\newblock 2009.
\newblock Domain adaptation from multiple sources via auxiliary classifiers.
\newblock In {\em Proceedings of the 26th Annual International Conference on
  Machine Learning},  289--296.
\newblock ACM.

\bibitem[\protect\citeauthoryear{Duan, Xu, and Tsang}{2012}]{duan2012domain}
Duan, L.; Xu, D.; and Tsang, I. W.-H.
\newblock 2012.
\newblock Domain adaptation from multiple sources: A domain-dependent
  regularization approach.
\newblock {\em IEEE Transactions on Neural Networks and Learning Systems}
  23(3):504--518.

\bibitem[\protect\citeauthoryear{Fang \bgroup et al\mbox.\egroup
  }{2018}]{FangXu2018Dart}
Fang, X.; Bai, H.; Guo, Z.; Shen, B.; Hoi, S. C.~H.; and Xu, Z.
\newblock 2018.
\newblock {DART:} domain-adversarial residual-transfer networks for
  unsupervised cross-domain image classification.
\newblock {\em CoRR} abs/1812.11478.

\bibitem[\protect\citeauthoryear{Ganin and
  Lempitsky}{2014}]{ganin2014unsupervised}
Ganin, Y., and Lempitsky, V.
\newblock 2014.
\newblock Unsupervised domain adaptation by backpropagation.
\newblock {\em arXiv preprint arXiv:1409.7495}.

\bibitem[\protect\citeauthoryear{Ganin \bgroup et al\mbox.\egroup
  }{2016}]{ganin2016domain}
Ganin, Y.; Ustinova, E.; Ajakan, H.; Germain, P.; Larochelle, H.; Laviolette,
  F.; Marchand, M.; and Lempitsky, V.
\newblock 2016.
\newblock Domain-adversarial training of neural networks.
\newblock {\em The Journal of Machine Learning Research} 17(1):2096--2030.

\bibitem[\protect\citeauthoryear{Gholami \bgroup et al\mbox.\egroup
  }{2018}]{gholami2018unsupervised}
Gholami, B.; Sahu, P.; Rudovic, O.; Bousmalis, K.; and Pavlovic, V.
\newblock 2018.
\newblock Unsupervised multi-target domain adaptation: An information theoretic
  approach.
\newblock {\em arXiv preprint arXiv:1810.11547}.

\bibitem[\protect\citeauthoryear{Johnson and
  Zhang}{2014}]{johnson2014effective}
Johnson, R., and Zhang, T.
\newblock 2014.
\newblock Effective use of word order for text categorization with
  convolutional neural networks.
\newblock {\em arXiv preprint arXiv:1412.1058}.

\bibitem[\protect\citeauthoryear{Le and Mikolov}{2014}]{le2014distributed}
Le, Q., and Mikolov, T.
\newblock 2014.
\newblock Distributed representations of sentences and documents.
\newblock In {\em International conference on machine learning},  1188--1196.

\bibitem[\protect\citeauthoryear{Liu, Qiu, and
  Huang}{2017}]{liu2017adversarial}
Liu, P.; Qiu, X.; and Huang, X.
\newblock 2017.
\newblock Adversarial multi-task learning for text classification.
\newblock {\em arXiv preprint arXiv:1704.05742}.

\bibitem[\protect\citeauthoryear{Liu}{2015}]{liu2015sentiment}
Liu, B.
\newblock 2015.
\newblock {\em Sentiment analysis: Mining opinions, sentiments, and emotions}.
\newblock Cambridge University Press.

\bibitem[\protect\citeauthoryear{Mikolov \bgroup et al\mbox.\egroup
  }{2013}]{mikolov2013efficient}
Mikolov, T.; Chen, K.; Corrado, G.; and Dean, J.
\newblock 2013.
\newblock Efficient estimation of word representations in vector space.
\newblock {\em arXiv preprint arXiv:1301.3781}.

\bibitem[\protect\citeauthoryear{Moraes, Valiati, and
  Neto}{2013}]{moraes2013document}
Moraes, R.; Valiati, J.~F.; and Neto, W. P.~G.
\newblock 2013.
\newblock Document-level sentiment classification: An empirical comparison
  between svm and ann.
\newblock {\em Expert Systems with Applications} 40(2):621--633.

\bibitem[\protect\citeauthoryear{Saito, Ushiku, and
  Harada}{2017}]{saito2017asymmetric}
Saito, K.; Ushiku, Y.; and Harada, T.
\newblock 2017.
\newblock Asymmetric tri-training for unsupervised domain adaptation.
\newblock In {\em Proceedings of the 34th International Conference on Machine
  Learning-Volume 70},  2988--2997.
\newblock JMLR. org.

\bibitem[\protect\citeauthoryear{Salzmann \bgroup et al\mbox.\egroup
  }{2010}]{salzmann2010factorized}
Salzmann, M.; Ek, C.~H.; Urtasun, R.; and Darrell, T.
\newblock 2010.
\newblock Factorized orthogonal latent spaces.
\newblock In {\em Proceedings of the Thirteenth International Conference on
  Artificial Intelligence and Statistics},  701--708.

\bibitem[\protect\citeauthoryear{Socher \bgroup et al\mbox.\egroup
  }{2013}]{socher2013recursive}
Socher, R.; Perelygin, A.; Wu, J.; Chuang, J.; Manning, C.~D.; Ng, A.; and
  Potts, C.
\newblock 2013.
\newblock Recursive deep models for semantic compositionality over a sentiment
  treebank.
\newblock In {\em Proceedings of the 2013 conference on empirical methods in
  natural language processing},  1631--1642.

\bibitem[\protect\citeauthoryear{Sun \bgroup et al\mbox.\egroup
  }{2011}]{sun2011two}
Sun, Q.; Chattopadhyay, R.; Panchanathan, S.; and Ye, J.
\newblock 2011.
\newblock A two-stage weighting framework for multi-source domain adaptation.
\newblock In {\em Advances in neural information processing systems},
  505--513.

\bibitem[\protect\citeauthoryear{Sun, Shi, and Wu}{2015}]{sun2015survey}
Sun, S.; Shi, H.; and Wu, Y.
\newblock 2015.
\newblock A survey of multi-source domain adaptation.
\newblock {\em Information Fusion} 24:84--92.

\bibitem[\protect\citeauthoryear{Zhao \bgroup et al\mbox.\egroup
  }{2017}]{zhao2017multiple}
Zhao, H.; Zhang, S.; Wu, G.; Costeira, J.~P.; Moura, J.~M.; and Gordon, G.~J.
\newblock 2017.
\newblock Multiple source domain adaptation with adversarial training of neural
  networks.
\newblock {\em arXiv preprint arXiv:1705.09684}.

\bibitem[\protect\citeauthoryear{Zhao \bgroup et al\mbox.\egroup
  }{2018}]{zhao2018multiple}
Zhao, H.; Zhang, S.; Wu, G.; Gordon, G.~J.; et~al.
\newblock 2018.
\newblock Multiple source domain adaptation with adversarial learning.

\bibitem[\protect\citeauthoryear{Zhou and Li}{2005}]{zhou2005tri}
Zhou, Z.-H., and Li, M.
\newblock 2005.
\newblock Tri-training: Exploiting unlabeled data using three classifiers.
\newblock {\em IEEE Transactions on Knowledge \& Data Engineering}
  (11):1529--1541.

\bibitem[\protect\citeauthoryear{Zhou, Wan, and Xiao}{2016}]{zhou2016attention}
Zhou, X.; Wan, X.; and Xiao, J.
\newblock 2016.
\newblock Attention-based lstm network for cross-lingual sentiment
  classification.
\newblock In {\em Proceedings of the 2016 conference on empirical methods in
  natural language processing},  247--256.

\end{thebibliography}

\bibliographystyle{aaai}
\bigskip

\end{document}